\documentclass[twoside,leqno,twocolumn]{article}

\usepackage[letterpaper]{geometry}

\usepackage{ltexpprt}
\usepackage{hyperref}

\usepackage[utf8]{inputenc} 
\usepackage[T1]{fontenc}    
\usepackage{hyperref}       
\usepackage{url}            
\usepackage{booktabs}       
\usepackage{amsfonts}       
\usepackage{nicefrac}       
\usepackage{microtype}      
\usepackage{graphicx}
\usepackage{multirow}
\usepackage{subcaption}
\usepackage{amsmath}
\usepackage{amssymb}
\usepackage{bbm}
\usepackage{enumitem}
\setlist{leftmargin=3.5mm}

\usepackage{bbding} 
\newtheorem{definition}{Definition}[section]
\usepackage[table]{xcolor}

\begin{document}

\title{PDT: \underline{P}retrained \underline{D}ual \underline{T}ransformers for \\ Time-aware Bipartite Graphs}

\author{
  Xin Dai{\Envelope} \thanks{These authors contributed equally to this work. Correspondence email: \texttt{xidai@visa.com}} \footnotemark[2]  \and
  Yujie Fan\footnotemark[1] \footnotemark[2]  \and
  Zhongfang Zhuang\footnotemark[1] \footnotemark[2] \and
  Shubham Jain\thanks{Visa Research, Palo Alto, CA.}\and
  Michael Yeh\footnotemark[2]\and
  Junpeng Wang\footnotemark[2]\and
  Liang Wang\footnotemark[2]\and
  Yan Zheng\footnotemark[2]\and
  Prince Aboagye\footnotemark[2]\and
  Wei Zhang\footnotemark[1] \footnotemark[2]
}

\date{}
\maketitle
\begin{abstract}
\textit{Pre-training} on large models is prevalent and emerging with the ever-growing user-generated content in many machine learning application categories. 
It has been recognized that learning contextual knowledge from the datasets depicting user-content interaction plays a vital role in downstream tasks. 
Despite several studies attempting to learn contextual knowledge via pre-training methods, finding an optimal training objective and strategy for this type of task remains a challenging problem. 
In this work, we contend that there are two distinct aspects of contextual knowledge, namely the \textit{user-side} and the \textit{content-side}, for datasets where user-content interaction can be represented as a bipartite graph. 
To learn contextual knowledge, we propose a pre-training method that learns a bi-directional mapping between the spaces of the user-side and the content-side. 
We formulate the training goal as a contrastive learning task and propose a dual-Transformer architecture to encode the contextual knowledge. 
We evaluate the proposed method for the recommendation task. 
The empirical studies have demonstrated that the proposed method outperformed all the baselines with up to 50.19\% in NDCG and 23.92\% in Recall@10 improvement. 
\end{abstract}


\section{Introduction}
\label{sec:intro}
Human activities, online and offline, have generated vast amounts of data representing user-content interactions. For example, we purchase at a coffee shop, place Amazon orders, or rate movies on IMDb. We refer to this type of data as user-content interaction data, where the contents can be movies, products, merchants, news feeds or ads. Data mining applications using these types of data can generate significant business benefits. For example, Netflix needs to make personalized recommendations to their subscribers, and Facebook needs to send proper feeds and ads to their users. 

Fundamentally, a common \textit{goal} of data mining applications using user-content interactions is to understand \textit{user's behaviors}~\cite{zhu2021contrastive} and content's properties. Researchers attempt multiple ways to model such behaviors: The \textit{time}-related nature of the interactions is a fit for sequential models, such as recurrent neural networks (RNN), and the interactions and relations can be modeled as graph neural networks (GNN). Conventionally, the training objective is to minimize the loss of a specific task such that one model is tailored to a particular application (e.g., recommendation). This approach is simple and effective for every data mining application. 

However, researchers have observed that the models trained solely to fit a specific task's objective may suffer from the issue of data sparsity, which may ultimately limit the model performance~\cite{zhou2020s3}. 
A feasible solution to this problem is \textit{self-supervised learning}, which aims to learn the contextual knowledge in an unsupervised fashion from the information, without relying on the supervised signal of specific data mining tasks, and \textit{refine} such models for specific downstream tasks. 
The current studies of self-supervised learning on tabular sequence data primarily focus on learning the contextual knowledge on the user-side, e.g., distinguishing positive samples and negative samples of a user's profile or predicting the masked items in a user's activity sequence~\cite{zhou2020s3, zhu2021contrastive, xie2022contrastive}. Even though they have shown success in empirical evaluation, their basic ideas are fundamentally inherited from the areas of computer vision and natural language modeling. 

However, in the user-content relationship, the context of a user's activity is not limited to their personal history, which differs from images or text, where the context is inherent in the image or text itself. For example, contents in a user's activity history may have also been purchased by other users. This context of the item side is not explicitly encoded in one user's activity.
Therefore, the unique and critical challenge for self-supervised learning on user-content datasets is identifying meaningful context.

In this paper, we formulate the user-content interactions as a bipartite graph. Then we can see two informative contexts closely related to each other: the activities on the user and content-sides. The pre-training task can be defined as two goals: (1) matching the user embedding to the encoded context from the user-side; (2) matching the content embedding to the encoded context from the content-side. Both of these goals can be formulated as contrastive learning tasks. The embeddings and context encoders can be jointly trained by back-propagation and be used for downstream tasks. We further proposed a dual Transformers architecture for pre-training and used the sequential recommendation as the downstream task. 

We evaluate the proposed method on the two large datasets: 1) MovieLens 25M, which consists of 25 million samples; 2) Amazon review dataset, which consists of 192 million samples. We compare the proposed approach to conventional and sequential recommendation models, including the GNNs and Transformers. We also compared with pre-training methods, e.g., Our results show that the proposed method outperformed all the baselines significantly. 
Moreover, we did an in-depth ablation study, showing how the proposed pre-training method contributes to the overall performance. We also visualized the embeddings learned from pre-training, demonstrating that the proposed method can discover the informative structure from the user-content two contexts mentioned above.

\begin{figure}[t]
\centering
\begin{subfigure}[c]{0.4\textwidth}
    \centering
    \includegraphics[width=0.3\textwidth]{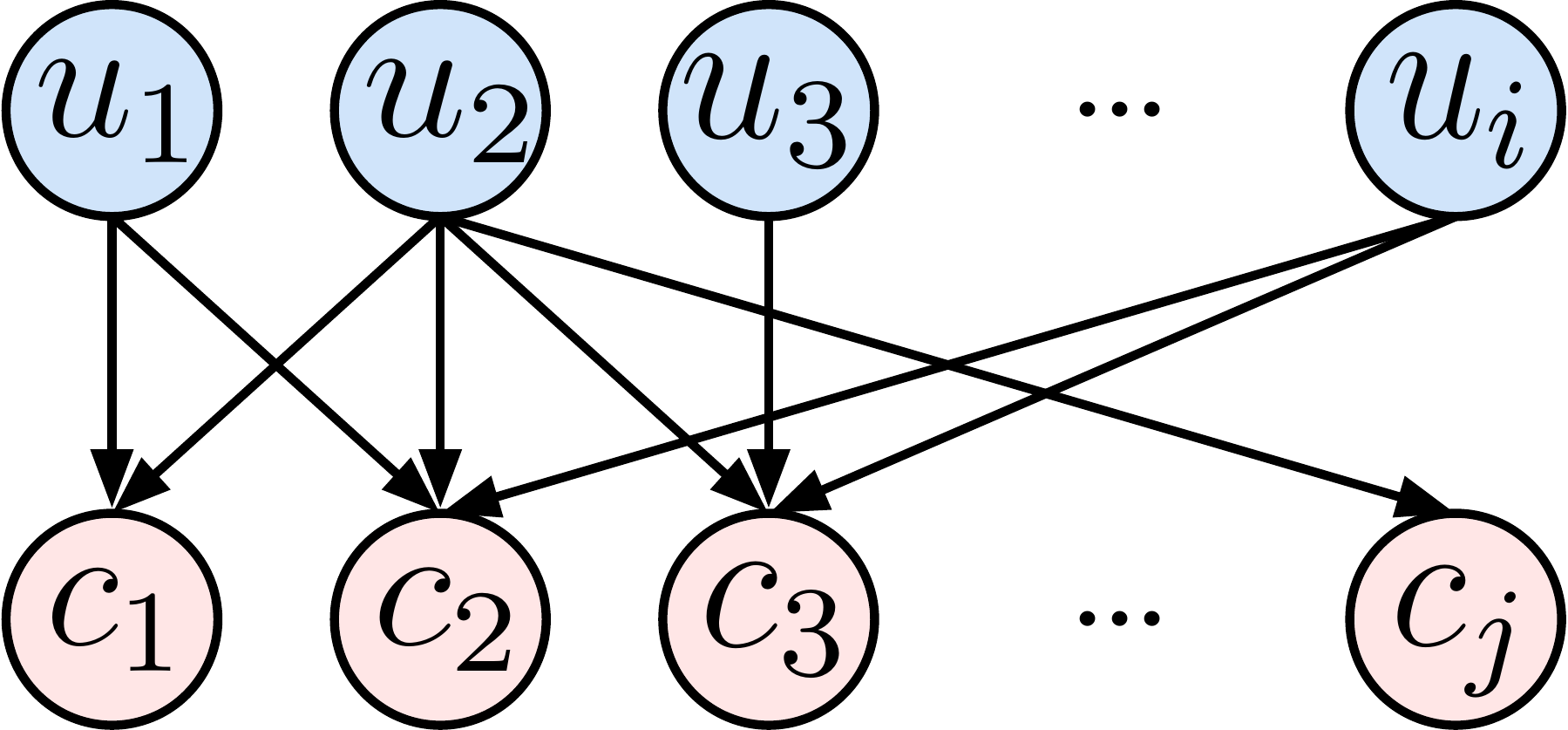}
    \caption{User-content dataset.}
    \label{fig-uc-db}
\end{subfigure}

\begin{subfigure}[c]{0.4\linewidth}
    \centering
    \includegraphics[width=0.7\textwidth]{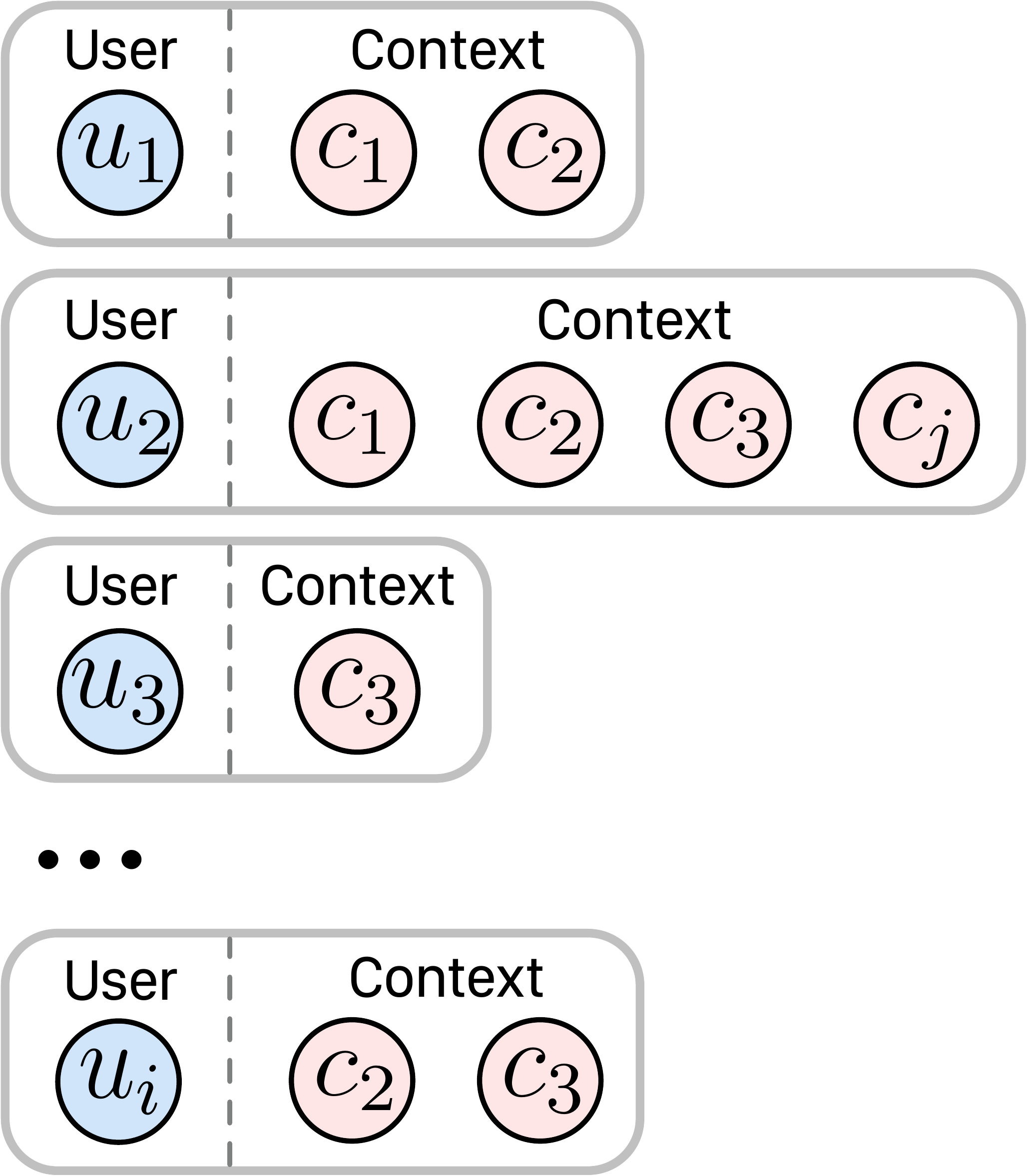}
    \caption{User-side context. }
    \label{fig-user-context}
\end{subfigure}
\begin{subfigure}[c]{0.4\linewidth}
    \centering
    \includegraphics[width=0.55\textwidth]{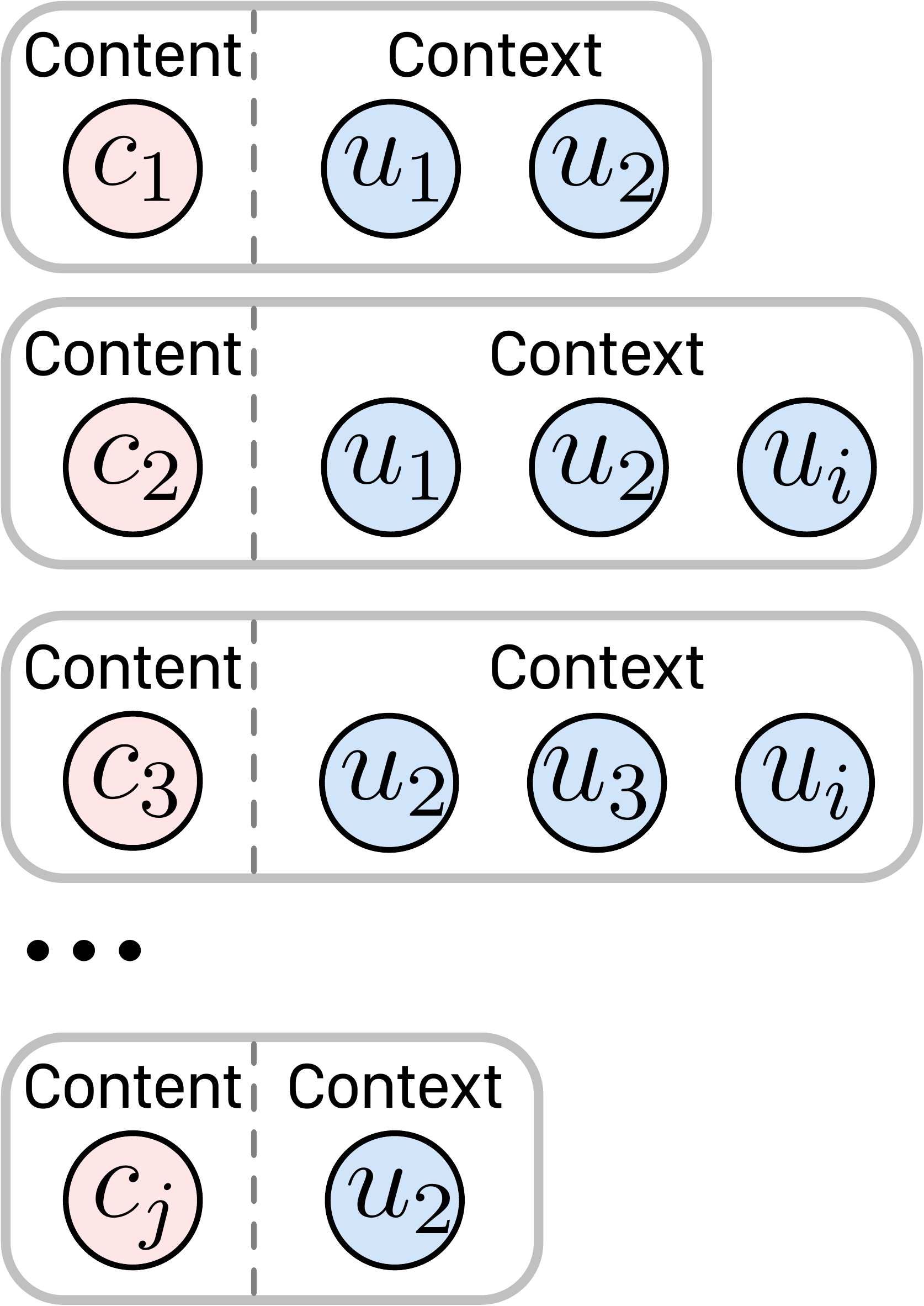}
    \caption{Content-side context. }
    \label{fig-content-context}
\end{subfigure}
\caption{An example of a user-content dataset in the form of a bipartite graph. }
\end{figure}


\section{Problem Formulation} 
We define a dataset as a bipartite graph $\mathcal{G} = \left\{\mathcal{U}, \mathcal{C}, \mathcal{E}\right\}$, where $\mathcal{U}$ is the vertex set of \textit{users}, $\mathcal{C}$ is the vertex set of \textit{contents}, in which each user is defined as $u_i \in \mathcal{U}$ and each content is defined as $c_j \in \mathcal{C}$. We also define $\mathcal{E}$ as the \textit{interaction} set, where each edge $e_{i,j}=\left(u_i, c_j\right) \in \mathcal{E}$ represents an interaction between a user $u_i$ and a content $c_j$.
We further define two types of context in a dataset:
\begin{definition}
    The \textit{user behavior history} $U_{i}$. Given a user $u_i$, the historical behavior of user $u_i$ can be denoted as a vector of edges: $U_{i} = \left(e_{i1}, \cdots, e_{in}\right)$, where $c_1, \cdots, c_n$ are the corresponding contents that have interactions with $u_i$ through the edges. In a general setting, $e_{1j}$ can be a multivariate vector if it has attributes. 
\end{definition}
\begin{definition}
    The \textit{content history} $C_{j}$. Giver a \textit{content} $c_j$, the content's history can be denoted as a vector $C_{j} = \left(e_{1j}, \cdots, e_{nj}\right)$, where $u_1, \cdots, u_n$ are the users that have interactions with $c_j$.
\end{definition}
To simplify the notation, we denote $U_{i}$ as $\left(c_1, \cdots, c_n\right)$ when there are no attributes associated with each edge $e_{ij}$, and $C_{j}$ can be simply denoted as $\left(u_1, \cdots, u_n\right)$ if $c_j$ is univariate. 

We further define the concept of heterogeneous representation of entities as follows: 
\begin{definition}
    Given the embedding of a user $u_i$ or a content $c_j$, denoted as $\texttt{emb}\left({u_i}\right)$ and $\texttt{emb}\left({c_j}\right)$, the \textit{context} of the user $u_i$ and the content $c_j$, in the form of the user behavior history $U_i$ and content history $C_j$, are the heterogeneous representation of the same \textit{entity}, respectively. 
\end{definition}

We can get a bi-directional mapping between the spaces of users and contents via \textit{learning to match} the two representations. 
Therefore, we define the goal of pre-training as jointly maximizing the mutual information $I\bigl(U_i;\texttt{emb}\left({u_i}\right)\bigr)$ and $I\bigl(C_j;\texttt{emb}\left({c_j}\right)\bigr)$. Specifically, our objective is to maximize the following: 
\begin{equation}
    I = I\bigl( U_i;\texttt{emb}\left({u_i}\right)\bigr) + I\bigl(C_j;\texttt{emb}\left({c_j}\right)\bigr)
\end{equation}
where the mutual information $I$ is defined as:
\begin{equation}
I(X; Y) = \sum_{y\in Y} \sum_{x\in X} P(x,y) \log\frac{P(x,y)}{P(x)P(y)}
\end{equation}
Since it is intractable to optimize this objective directly,~\cite{oord2018representation} has proved that minimizing \texttt{InfoNCE} loss maximizes a lower bound on the mutual information. Therefore, our training objective is to minimize the \texttt{InfoNCE} loss.

\begin{figure*}
  \centering
    \begin{minipage}[c]{\textwidth}
        \centering
		\includegraphics[width=\textwidth]{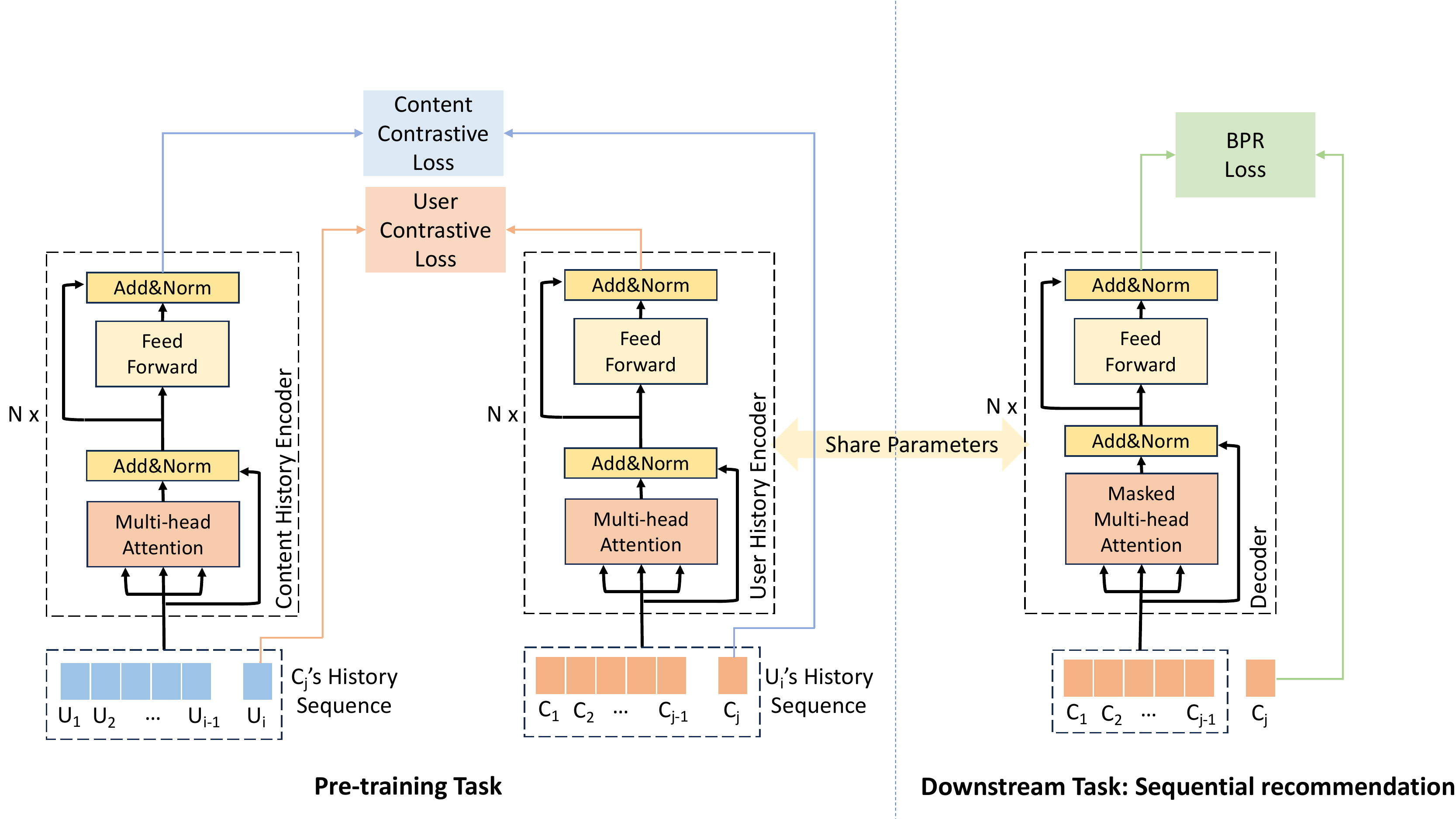}
    \end{minipage}
\caption{\textbf{ The model architecture of PDT}. 
}
\vspace{-15pt}
\label{fig:mothod_pretrain}
\end{figure*}

\section{The Proposed PDT Model}
\label{sec:method}
In this section, we first present the model architecture for the pre-training model. After that, we elaborate on the downstream tasks using recommendation engines as an example. 
\subsection{Model Architecture for Pre-training.}
\label{sec:pretraining}
Given an edge $e_{ij}$ from a time-aware bipartite graph, we have the user $u_i$'s behavior history $U_i$ and content $c_j$'s history $C_j$ before the timestamp of $e_{ij}$. Our main objective is to learn how to map $U_i$ and the embedding of $u_i$ to proximate regions in a latent space and similarly to map $C_j$ and the embedding of $c_j$ to nearby regions in a separate latent space.

Our proposed model consists of four parts: (1) The user embedding layer $f_{u}$, (2) the content embedding layer $f_{c}$, (3) the user history encoder $g_u$ and (4) the content history encoder $g_c$. Similar with SASRec model~\cite{SASRec}, we perform layrnorm and dropout on user and content embeddings before the transformer encoders.

\textbf{User history encoder $g_u$} is a multi-head Transformer encoder. Given a user $u_i$, It takes $u_i$'s behavior history $U_i = \bigl(f_c(c_1), \cdots, f_c(c_j)\bigr)$ as input and outputs its \texttt{CLS} token as an encoded representation of $U_i$. Here we denote the output as $g_u(U_i)$.

\textbf{Content history encoder $g_c$} is also a multi-head Transformer encoder. Given a content $c_j$, It takes $c_j$'s behavior history $C_j = \bigl(f_u(u_1), \cdots, f_u(u_i)\bigr)$ as input and outputs its \texttt{CLS} token as an encoded representation of $C_j$. Here we denote the output as $g_c(C_j)$. 

Given an edge $e_{ij}$ we define the contrastive loss between the user behavior $U_i$  and user embedding $f_u(u_i)$ as:

\begin{equation}
\mathcal{L}_{u} = - \log \frac{\exp \biggl[\mathrm{sim}\bigl(f_u(u_i), g_u(U_i)\bigr)\biggr]}{\sum_{k=1}^{N} \mathbbm{1}_{[k \neq i]} \exp\biggl[\mathrm{sim}\bigl(f_u(u_k), g_u(U_i)\bigr)\biggr]}
\end{equation}

Here $\mathrm{sim}$ function is inner product. In the same way, For the content $c_j$ we define the contrastive loss between the content history $C_j$ and content embedding $f_c(c_j)$ as:

\begin{equation}
\mathcal{L}_{c} = - \log \frac{\exp \biggl[\mathrm{sim}\bigl(f_c(c_j), g_c(C_j)\bigr)\biggr]}{\sum_{k=1}^{N} \mathbbm{1}_{[k \neq i]} \exp\biggl[\mathrm{sim}\bigl(f_c(c_k), g_c(C_j)\bigr)\biggr]}
\end{equation}


\subsection{Model Architecture for Recommendation Task.}

This paper takes the sequential recommendation as an example of a downstream task to test our proposed method. The task is formulated as given a user $u_i$'s recent behavior history $U_i$, predicting the next item $c_\text{next}$ that $u_i$ may interact with. We use a decoder-only architecture for this task. The predicted item is calculated as:

\begin{equation}
y_i = g_r\bigl(U_i\bigr)
\end{equation}

Here $g_r$ is a transformer decoder, and we adopt the SASRec model as its backbone. The decoder $g_r$, the encoders $g_u$ and $g_c$ have the same architecture, except for two differences:

1) $g_r$ doesn't have \texttt{CLS} token. It outputs the final representation of the last token in $U_i$ as the predicted item $y_i$.

2) $g_r$ utilizes unidirectional self-attention, whereas $g_u$ and $g_c$ employ bidirectional self-attention. This distinction is why we refer to $g_r$ as a decoder, while considering $g_u$ and $g_c$ as encoders.

The decoder $g_r$ and encoder $g_u$ share all parameters i.e. weights and biases, except for the \texttt{CLS} token in $g_u$. While there are no shared parameters between $g_c$ and $g_r$, the former indirectly influences the latter through its impact on training embeddings.

Similar with pre-training, the inner product $y_i^T \cdot f_c(c)$ is adopted to measure how likely a user $u_i$ would interact with item $c$. We jointly use the contrastive losses $\mathcal{L}_{u}$, $\mathcal{L}_{c}$ and BPR loss to fine-tune the Recommendation task:
\begin{equation}
\mathcal{L}_\text{BPR} = - \log \biggl[\sigma\bigl(y_i^{\intercal} \cdot f_c(c_\text{pos})\bigr) - \sigma\bigl(y_i^\intercal \cdot f_c(c_\text{neg})\bigr)\biggr]
\end{equation}
\begin{equation}
\mathcal{L}_\text{total} = \mathcal{L}_\text{BPR} + \lambda_u \cdot\mathcal{L}_{u_i} +  \lambda_c \cdot\mathcal{L}_{c_j}
\end{equation}

Here $c_\text{pos}$ and $c_\text{neg}$ are the positive and negative samples of the next item, respectively. $\lambda_u$ and $\lambda_c$ are hyper-parameters controlling the weights of contrastive losses. 

\section{Experiments}

\subsection{Datasets.} 
We evaluate the proposed method on two large datasets:

(1) MovieLens-25M. The dataset consists of 25,000,095 reviews made by 162,541 reviewers on 59,047 movies. We performed the following prepossessing steps: First, group the reviews by reviewers and sort the reviews by timestamp. Second, for each reviewer, select the last review for testing, the second last for validation, and all remaining reviews for training. Third, we remove all the movies not in training set in validation and test sets because the cold start problem is beyond the paper's discussion. 

(2) Amazon review. The dataset consists of 192,464,864 reviews made by 47,249,276 reviewers on 14,894,121 products. We split the dataset into train, validation and test based on timestamp by 80\%, 10\%, 10\%. 

\subsection{Baselines.}

We evaluate our proposed PDT against two types of recommendation models: general recommenders and sequential recommenders. General recommenders rely on either matrix factorization (BPR-MF and NeuMF) or graph neural networks (NGCF and LightGCN). In contrast, sequential recommenders employ either convolutional neural networks (Caser), recurrent neural networks (GRU4Rec), or self-attention mechanisms (SASRec, GCSAN and SRGNN). Below are the baseline methods we use in this work:

\begin{itemize}
    \item \textbf{BPR-MF}~\cite{BPR} is a matrix factorization-based method that optimizes Bayesian personalized ranking (BPR) loss to learn user and item latent representations.
    \item \textbf{NeuMF}~\cite{NeuMF} uses multiple hidden layers above the element-wise and concatenation of user and item embeddings to capture their nonlinear feature interactions.
    \item \textbf{NGCF}~\cite{NGCF} is a graph-based collaborative filtering method that relies on standard GCN to capture high-order user-content interactions.
    \item \textbf{LightGCN}~\cite{Lightgcn} simplifies NGCF by disabling the feature transformation and nonlinear activation.
    \item \textbf{GRU4Rec}~\cite{GRU4Rec} applies gated recurrent units (GRUs) to model the user click sequence for the next item recommendation.
    \item \textbf{Caser}~\cite{Caser} leverages convolutional operations to capture the high-order Markov chains for sequence recommendation.
    \item \textbf{SASRec}~\cite{SASRec} is a self-attention-based sequential model which uses a multi-head attention mechanism to recommend the next item.
    \item \textbf{GCSAN}~\cite{GCSAN} applies graph neural networks (GNNs) and self-attention to capture local and long-range dependencies for session-based recommendation.
    \item \textbf{SRGNN}~\cite{SRGNN} also performs GNNs with attention network on the constructed session graph for session-based recommendation.
\end{itemize}

\subsection{Configuration.} 
\textbf{Network architecture:}
We implement the user and content behavior encoders as Transformer encoders of the same architecture. We set two layers and two heads in each encoder and the dimension of feed forward layer is 256, which is exactly same with~\cite{SASRec}. The dropout rate is 0.2. We use GeLU as the activation function. For Movielens dataset, we try two settings of dimension of user and item embedding: The version named PDT-Large uses 512 dimension, and the version named PDT-Samll uses 128 dimension. We set $\lambda_u$=$\lambda_c$=0.01. For Amazon review dataset, we use 128 dimension for product item embedding, and 8 dimension for user embedding, due to the huge amount of items and users. To solve the dimension miss matching between items and users, we use a linear transformation to project user embeddings to 128 dimension. The number of trainable parameters reaches to 2.2 billions.  We set $\lambda_u$=0.01 and $\lambda_c$=0.05.

\textbf{Hyperparameter for training:}
We use the Adam optimizer for training. The learning rate is 0.0001. The batch size is 1024. For the contrastive learning loss, we set the temperature to 0.5. For pre-training, we use 8 NVIDIA A100 GPUs and train 2 epochs. Finally, we fine-tune the model by 30 epochs for the recommendation task. 

\newcolumntype{P}[1]{>{\centering\arraybackslash}p{#1}}

\begin{table*}[!h]
    \small
    \label{tab:comparison}
    \centering
    \caption{Comparison between PDT and Baselines on Movielens dataset. -Small and -Large indicate the hidden dimension $d=128$ and $d=512$, respectively. The best and second best results are highlighted in \textbf{bold} and \underline{underlined}. Improv. (SAS) and Improv. (Best) mean the performance improvement of PDT-Large over SASRec-Large and the best baseline.}
\begin{tabular}{c|c|P{1.4cm}P{1.4cm}P{1.4cm}P{1.4cm}P{1.4cm}P{1.4cm}}
\toprule
        & Model        & Recall@10 & Recall@20 & Recall@50 & NDCG@10 & NDCG@20 & NDCG@50 \\
\midrule
\multirow{4}{*}{\rotatebox[origin=c]{90}{General}}    
        & BPR-MF       & 0.0710    & 0.1215    & 0.2324    & 0.0349  & 0.0476  & 0.0695  \\
        & NeuMF        & 0.0778    & 0.1340    & 0.2524    & 0.0380  & 0.0521  & 0.0754  \\
        & NGCF         & 0.0712    & 0.1202    & 0.2282    & 0.0353  & 0.0476  & 0.0688  \\
        & LightGCN     & 0.0824    & 0.1381    & 0.2548    & 0.0412  & 0.0551  & 0.0781  \\
        \midrule
\multirow{6}{*}{\rotatebox[origin=c]{90}{Sequential}} 
        & GRU4Rec      & 0.1109    & 0.1839    & 0.3231    & 0.0527  & 0.0711  & 0.0985  \\
        & Caser        & 0.0929    & 0.1602    & 0.2953    & 0.0439  & 0.0608  & 0.0875  \\
        & SASRec-Small & 0.0940    & 0.1634    & 0.3033    & 0.0446  & 0.0620  & 0.0896  \\
        & SASRec-Large & 0.1179    & 0.1929    & 0.3337    & 0.0528  & 0.0716  & 0.0995  \\
        & GCSAN        & 0.1234    & 0.2014    & 0.3468    & 0.0593  & 0.0789  & 0.1077  \\
        & SRGNN        & 0.1261    & 0.1987    & 0.3316    & 0.0634  & 0.0817  & 0.1079  \\
        \midrule
        \rowcolor{gray!10}
        & PDT-Small    &\underline{0.1419}&\underline{0.2176}&\textbf{0.3566}&\underline{0.0749}&\underline{0.0939}&\underline{0.1214} \\
        \rowcolor{gray!10}
        & PDT-Large     &\textbf{0.1461}&\textbf{0.2206}&\underline{0.3559}&\textbf{0.0793}&\textbf{0.0980}&\textbf{0.1248} \\
        \rowcolor{gray!10}
        & Improv. (SAS)  & 23.92\% & 14.36\% & 6.65\% & 50.19\% & 36.87\% & 25.43\%    \\
        \rowcolor{gray!10}
        \multirow{-4}{*}{\rotatebox[origin=c]{90}{Ours}}
        & Improv. (Best)  & 15.86\%  & 9.53\%  & 2.62\%  & 25.08\%  & 19.95\% & 15.66\%     \\
        \midrule
        \rowcolor{gray!10}
        & w/o $\mathcal{L}_{u}$& 0.1448 & 0.2196 & 0.3534 & 0.0787 & 0.0975 & 0.1240       \\
        \rowcolor{gray!10}
        & w/o $\mathcal{L}_{c}$& 0.1395 & 0.2120 & 0.3476 & 0.0752 & 0.0935 & 0.1203         \\
        \rowcolor{gray!10}
        \multirow{-3}{*}{\rotatebox[origin=c]{90}{Ablation}}
        & w/o $\mathcal{L}_{u}+\mathcal{L}_{c}$ & 0.1179    & 0.1929    & 0.3337    & 0.0528  & 0.0716  & 0.0995         \\
        \bottomrule
\end{tabular}
\label{tab:results_movielens}
\end{table*}

\subsection{Evaluation Results on Movielens Dataset.} 

In the pre-training step, we set the history length to 9 for users and movies (the content-side). In the fine-tuning step, we set the history length to 8 for all baselines and our model. The final models for the test are selected based on their Recall@10 on the validation set.

We show the results of the recommendation task in Table~\ref{tab:results_movielens}. The results of sequential recommenders are generally better than the general recommenders because they take the user history as input. The proposed PDT models, including PDT-Large and PDT-Small, outperformed all baselines under all metrics. Compared with the performance of SASRec which is the backbone of our model, PDT achieved up to $50\%$ improvement. Also, PDT performs up to $25\%$ better than the best performances of other sequential recommenders.

\subsection{Evaluation Results on Amazon Review Dataset.} 
 
Due to the huge amount of products, evaluation by fully sorting the whole product set would be impossible. So for each user, we randomly select 10000 negative products which have not interacted with the user. The final models for the test are selected based on their Recall@10 on the validation set. The general recommenders work poorly on this dataset, and the graph models can't scale to such large dataset due to their high computational cost, so we only compared with SASRec and GRU4Rec.

We show the results of the recommendation task in Table~\ref{tab:results_amazon}. SASRec surpasses GRU4Rec, and our proposed PDT surpasses both of them across all metrics. The enhancements brought about by PDT become more pronounced as the value of K decreases for Recall@K and NDCG@K. We further considered different configurations of heads and layers. As the number of heads and layers increase from 2 to 32, the improvements compared with SASRec increase from $13.02\%$ to $28.34\%$ for Recall@10, and from $15.38\%$ to $33.49\%$ for NDCG@10. The performance of SASRec doesn't change a lot, or even decreases, as the model size increases, in contrast to PDT which demonstrates a clear upward trend. This observation underscores the efficacy of contrastive learning losses. By comparing the performance advantages gained by SASRec over GRU4Rec and by PDT over SASRec, we uncover a more interesting insight: the integration of the two contrastive loss brings about notably greater enhancements compared to the mere transition of the backbone architecture from RNN to Transformer.


\newcolumntype{P}[1]{>{\centering\arraybackslash}p{#1}}

\begin{table*}[!h]
    \small
    \label{tab:comparison}
    \centering
    \caption{Comparison between PDT and Baselines on Amazon review dataset. -($n\times n$) indicates \# heads = $n$ and \# layers = $n$. Improv.(SAS) and Ablation are for -($32\times 32$) models.}
\begin{tabular}{c|c|P{1.4cm}P{1.4cm}P{1.4cm}P{1.4cm}P{1.4cm}P{1.4cm}}
\toprule
        & Model                      & Recall@10 & Recall@20 & Recall@50 & NDCG@10 & NDCG@20 & NDCG@50\\
\midrule
\multirow{2}{*}{\rotatebox[origin=c]{90}{}} 
        & GRU4Rec                    & 0.0939    & 0.1409    & 0.2319    & 0.0527  & 0.0645  & 0.0825 \\
        & SASRec-($2\times 2$)       & 0.1113    & 0.1568    & 0.2409    & 0.0670  & 0.0788  & 0.0957 \\
        & SASRec-($8\times 8$)       & 0.1125    & 0.1594    & 0.2455    & 0.0676  & 0.0791  & 0.0958 \\
        & SASRec-($32\times 32$)     & 0.1069    & 0.1526    & 0.2375    & 0.0627  & 0.0742  & 0.0910 \\
        
        \midrule
        \rowcolor{gray!10}
        \rowcolor{gray!10}
        & PDT-($2\times 2$)          & 0.1258    & 0.1762    & 0.2635    & 0.0780  & 0.0912  & 0.1085 \\
        \rowcolor{gray!10}
        & PDT-($8\times 8$)          & \underline{0.1328}    & \underline{0.1834}    & \underline{0.2718}    & \underline{0.0819}  & \underline{0.0940}  & \underline{0.1122} \\
        \rowcolor{gray!10}
        & PDT-($32\times 32$)       & \textbf{0.1372}    & \textbf{0.1881}    & \textbf{0.2730}    & \textbf{0.0837}  & \textbf{0.0965}  & \textbf{0.1141}\\
        \rowcolor{gray!10}
        \multirow{-2}{*}{\rotatebox[origin=c]{90}{Ours}}
        & Improv. (SAS)  & 28.34\% & 23.26\% & 14.94\% & 33.49\% & 30.05\% & 25.38\% \\
        \rowcolor{gray!10}
        \midrule
        \rowcolor{gray!10}
        & w/o $\mathcal{L}_{u}$& 0.1307 & 0.1811 & 0.2696 & 0.0785 & 0.0912 & 0.1087 \\
        \rowcolor{gray!10}
        & w/o $\mathcal{L}_{c}$& 0.1091 & 0.1554 & 0.2398 & 0.0641 & 0.0757 & 0.0924 \\
        \rowcolor{gray!10}
        \multirow{-3}{*}{\rotatebox[origin=c]{90}{Ablation}}
        & w/o $\mathcal{L}_{u}+\mathcal{L}_{c}$ & 0.1069    & 0.1526    & 0.2375    & 0.0627  & 0.0742  & 0.0910 \\
        \bottomrule
\end{tabular}
\label{tab:results_amazon}
\end{table*}

\subsection{Ablation Study.} 
To investigate the contribution of each component in PDT, we designed three variants by disabling the corresponding loss function: 
\begin{itemize}
    \item  \textit{Without} $\mathcal{L}_{u}$: Disable the loss of user-side in both the pre-training and fine-tuning phases;
    \item  \textit{Without} $\mathcal{L}_{c}$: Disable the loss of content-side in both the pre-training and fine-tuning phases;
    \item  \textit{Without} $\mathcal{L}_{u}+\mathcal{L}_{c}$: Directly optimizes the model by $\mathcal{L}_\text{BPR}$ without pre-training. It is identical to SASRec.
\end{itemize}
These three variants are trained in the same configuration as the proposed method.
Their results of the recommendation task on Movielens and Amazon review dataset are shown in Table~\ref{tab:results_movielens} and Table~\ref{tab:results_amazon} respectively. The observations from the two datasets are surprisingly identical. We can see that w/o $\mathcal{L}_{u}+\mathcal{L}_{c}$ yielded the worst performance compared with the proposed method and other variants, indicating the importance and effectiveness of pre-training. The $\mathcal{L}_{u}$ and $\mathcal{L}_\text{BPR}$ can be both considered as modeling user behavior; however, w/o $\mathcal{L}_{c}$ outperformed w/o $\mathcal{L}_{u}+\mathcal{L}_{c}$, indicating that adding $\mathcal{L}_{u}$ term into the final loss function can learn more comprehensive user behavior than the one solely relying on $\mathcal{L}_\text{BPR}$. Additionally, w/o $\mathcal{L}_{u}$ outperformed w/o $\mathcal{L}_{c}$, which proves the effectiveness of the knowledge learned from content behavior. The proposed method outperformed w/o $\mathcal{L}_{c}$, confirming the effectiveness of the joint training by $\mathcal{L}_{u}$ and $\mathcal{L}_{c}$.

\subsection{Visualization and Case Study of Movie Embeddings.} 

To further analyze what the PDT learned from dataset, we conduct visualization and case study on Movielens dataset. We first normalize all movie embeddings by dividing by their l2 norms, then use tSNE to map the normalized embeddings to a 2D space to visualize the embeddings learned in the pre-training step. Then select the top 10,000 movies that received the most reviews to show. 

\begin{figure*}[t]
    \centering
    \hspace{-10pt}
    \begin{subfigure}{0.4\textwidth}
    \centering
        \includegraphics[width=\textwidth]{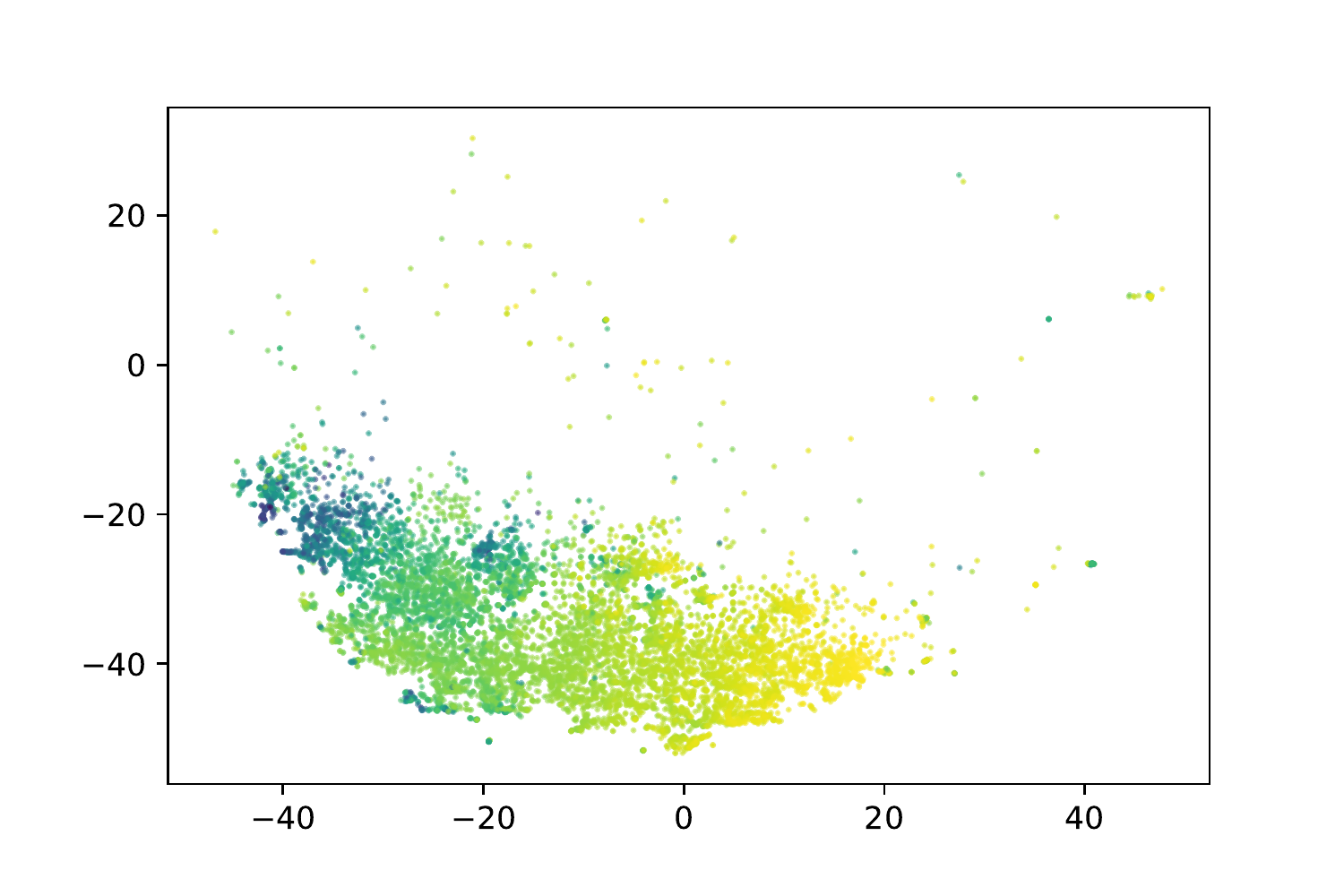}
        \caption{The releasing year(the greener the older)}
        \label{fig:movie_time}
    \end{subfigure}
    \vspace{-2pt}
    \hspace{10pt}
    \begin{subfigure}{0.4\textwidth}
    \centering
        \includegraphics[width=\textwidth]{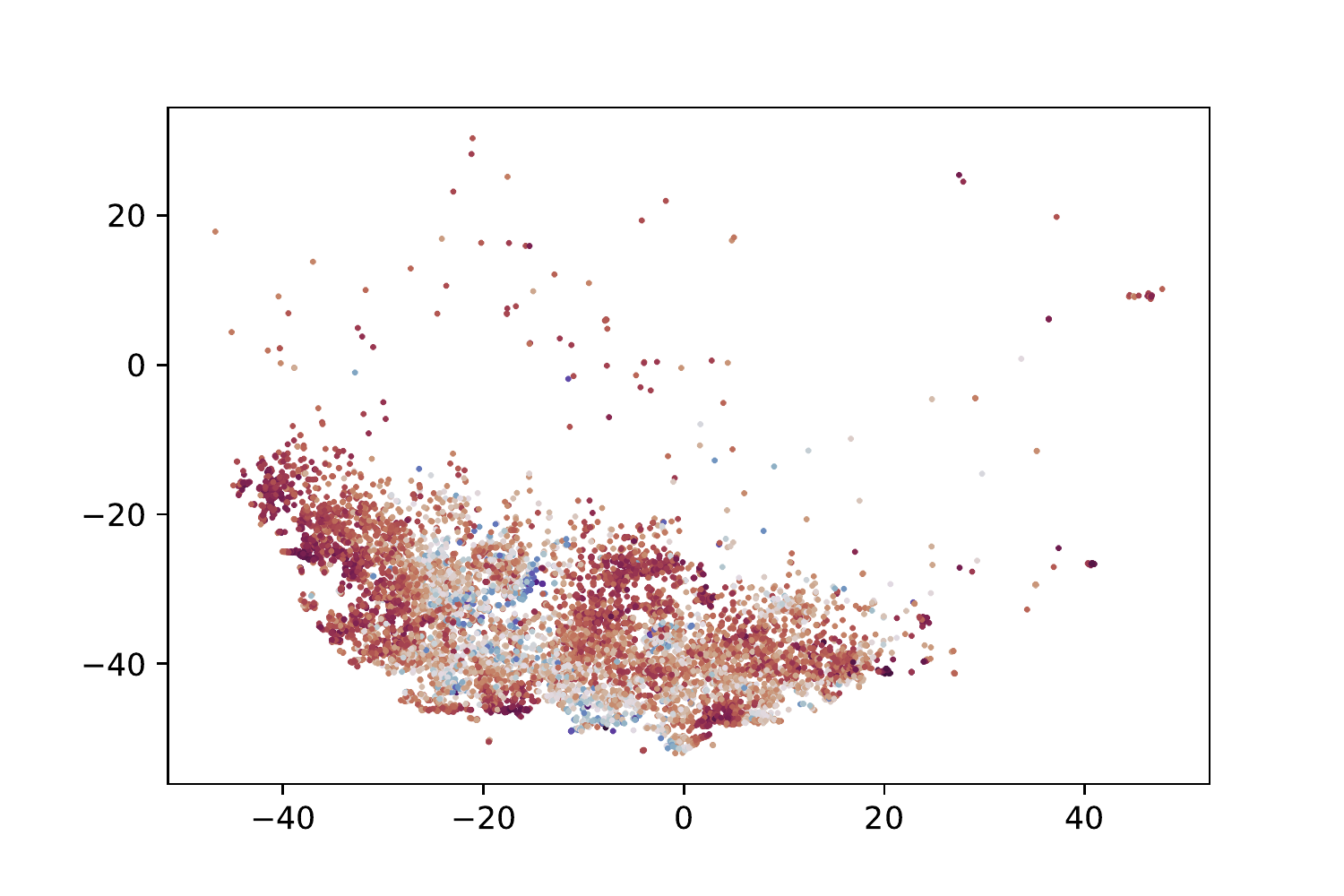}
        \caption{The average rating (the reder the higher)}
        \label{fig:movie_score}
    \end{subfigure}
\caption{\textbf{Visualizing movie embeddings in MovieLens 25M by \textit{releasing year} and \textit{avg. rating}.}}
\vspace{-10pt}
\label{fig:visual1}
\end{figure*}

The most significant patterns learned by the movie embeddings are the release year and average rating. In Figure~\ref{fig:visual1}, each dot represents a movie. Figure~\ref{fig:movie_time} shows the release year of movies. If the color appears more green, it indicates that the movie is from an earlier year. There is a clear trend that the green color is fading more and more from left to right. 
Figure~\ref{fig:movie_score} shows the average score of movies. The movie has a higher average rating if the color appears more red. The darker red dots are clustered in smaller regions at the corners of the big cluster. 

\begin{figure*}[t]
    \centering
    \begin{subfigure}{0.3\textwidth}
    \centering
        \includegraphics[width=\textwidth]{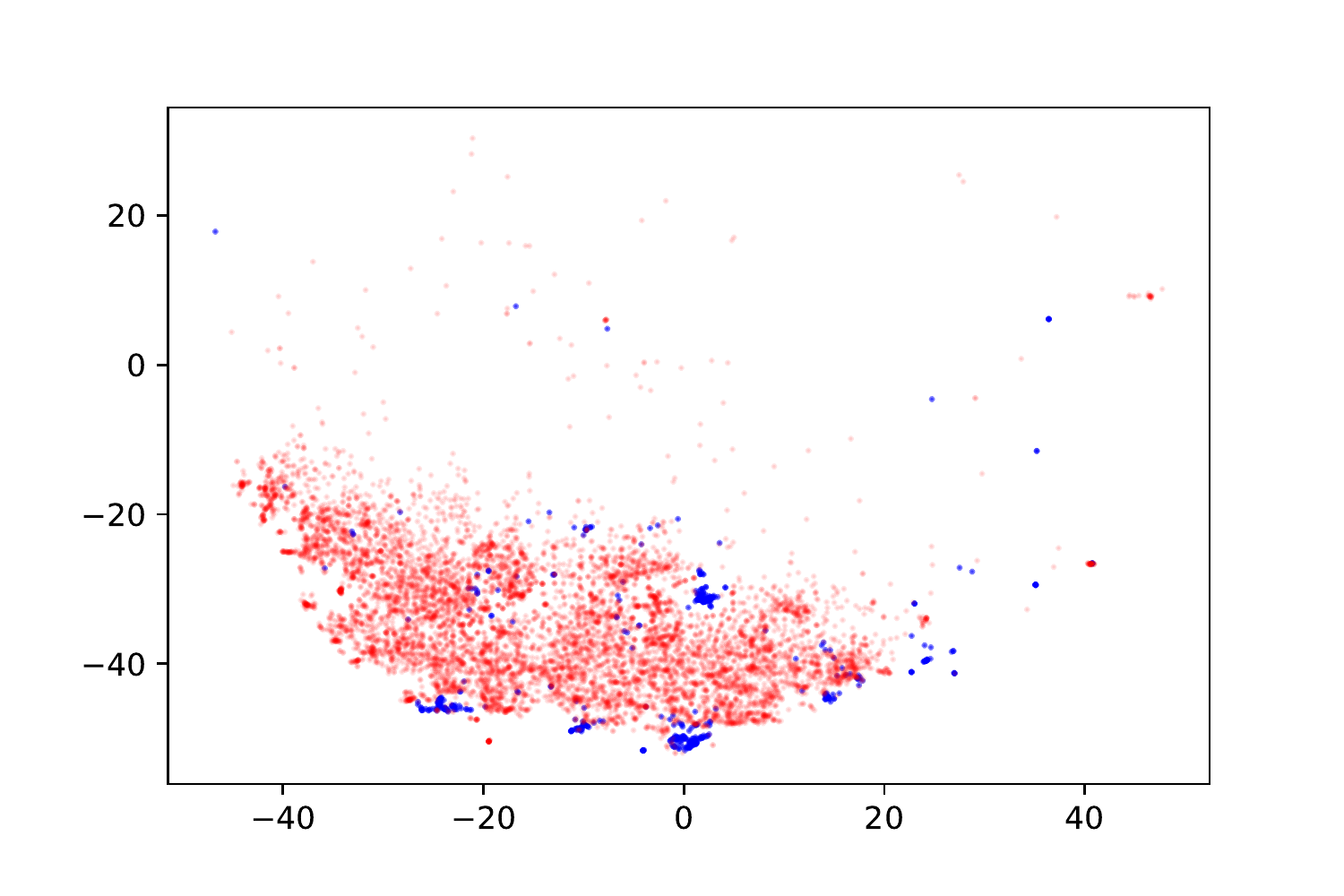}
        \caption{Animation}
        \label{fig:animation}
    \end{subfigure}
    \hspace{10pt}
    \begin{subfigure}{0.3\textwidth}
    \centering
        \includegraphics[width=\textwidth]{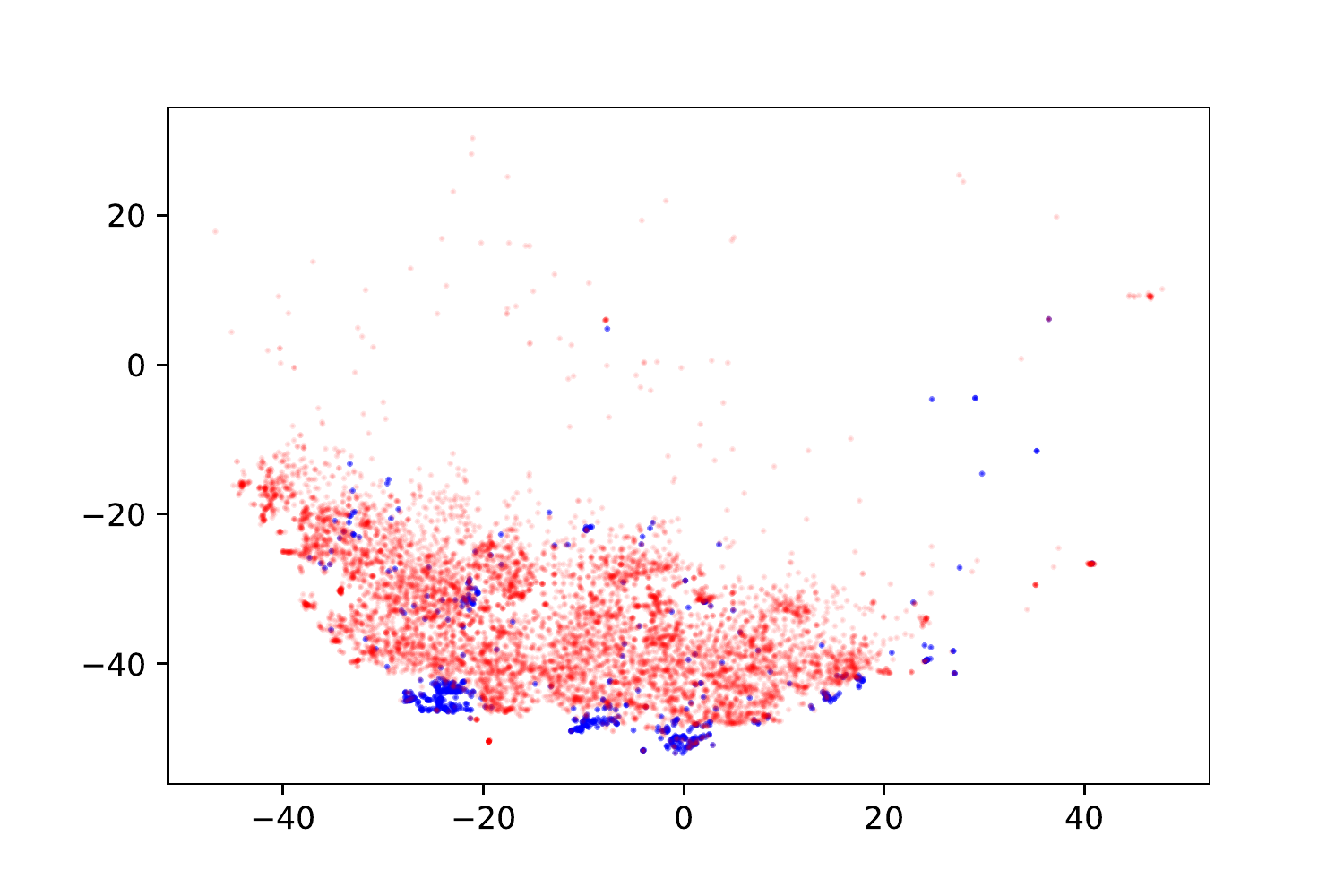}
        \caption{Children}
        \label{fig:children}
    \end{subfigure}
    \hspace{10pt}
    \begin{subfigure}{0.3\textwidth}
    \centering
        \includegraphics[width=\textwidth]{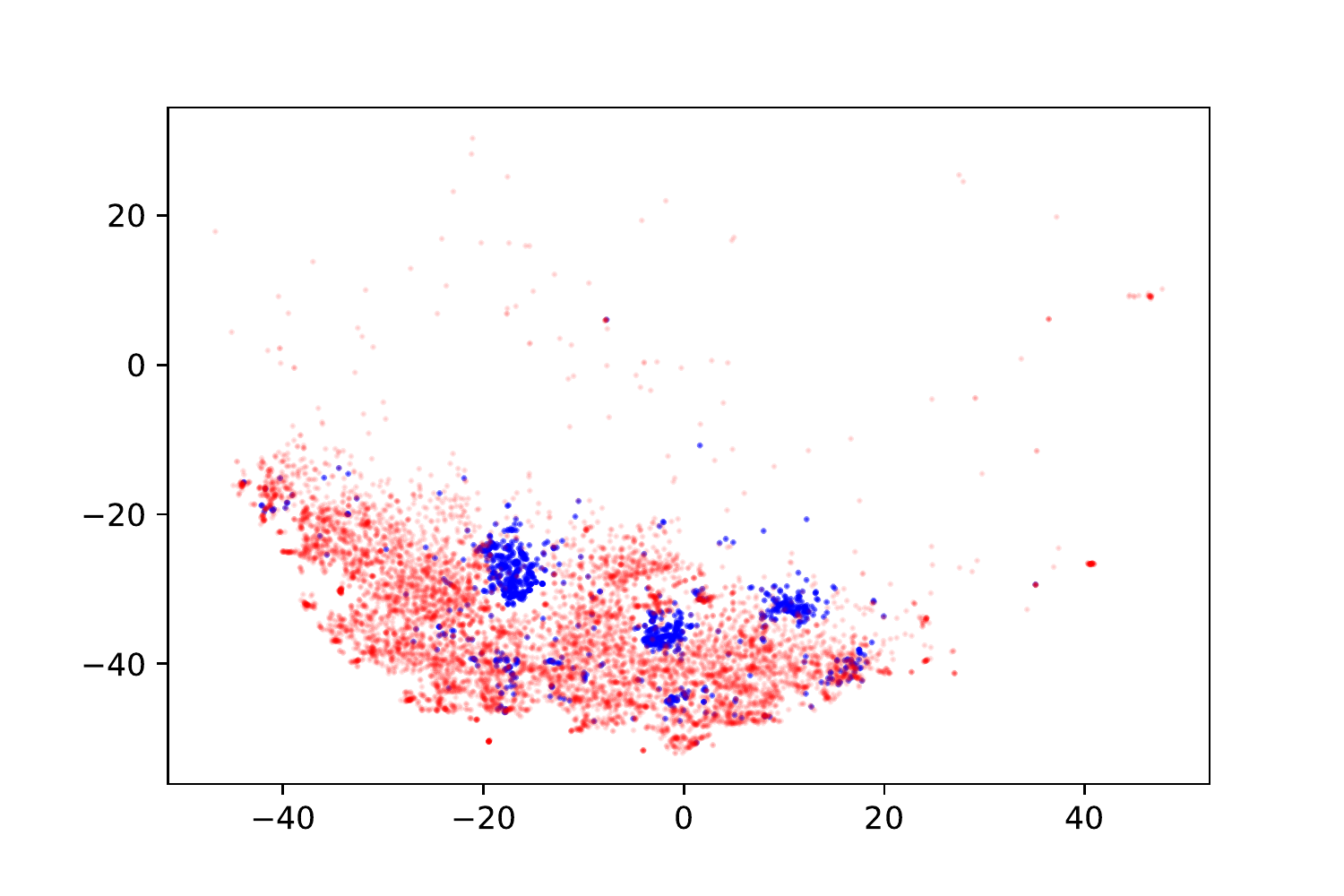}
        \caption{Horror}
        \label{fig:horror}
    \end{subfigure}
    \hspace{10pt}
    \begin{subfigure}{0.3\textwidth}
    \centering
        \includegraphics[width=\textwidth]{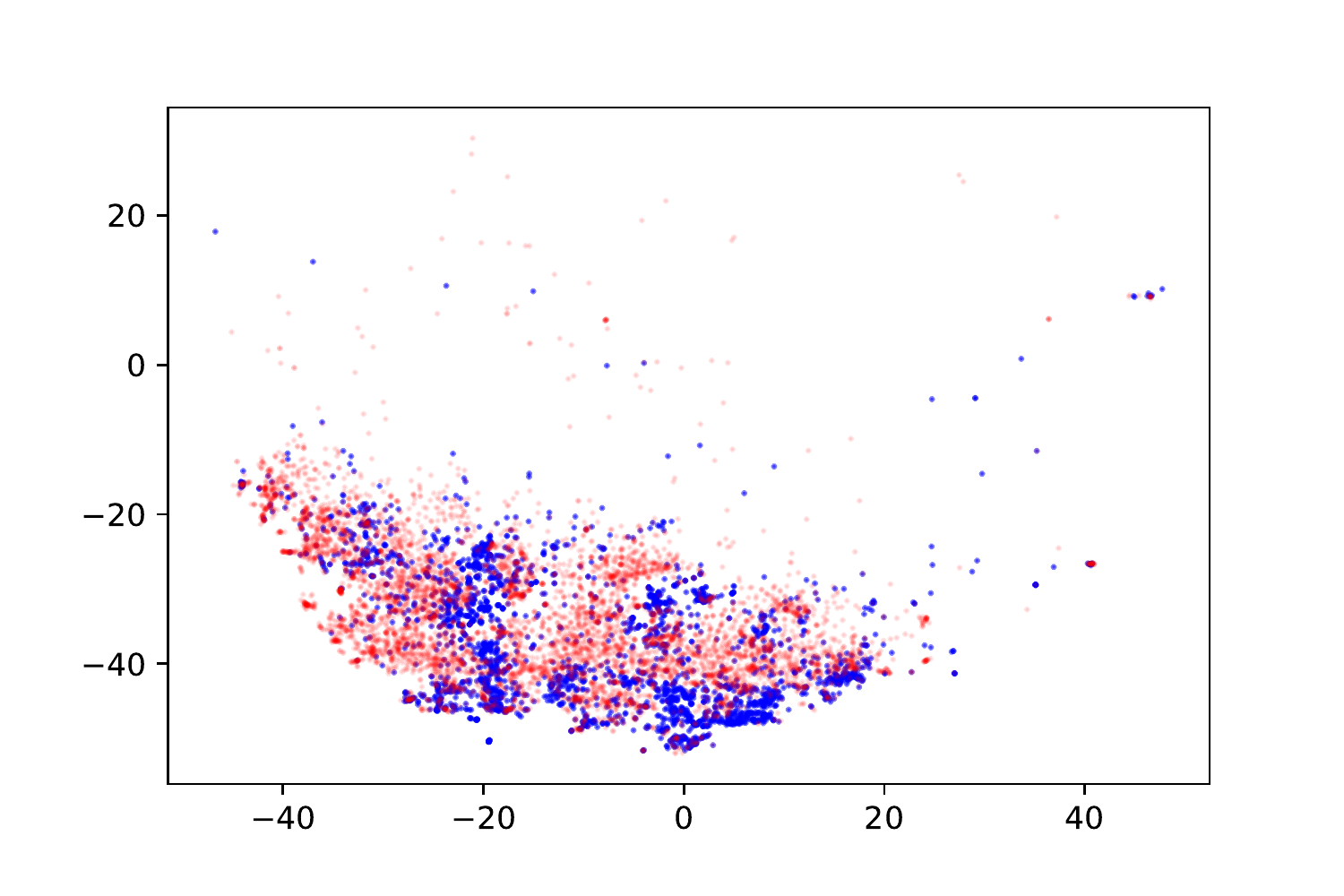}
        \caption{Action/Adventure/Sci-Fi}
        \label{fig:action}
    \end{subfigure}
    \hspace{10pt}
    \begin{subfigure}{0.3\textwidth}
    \centering
        \includegraphics[width=\textwidth]{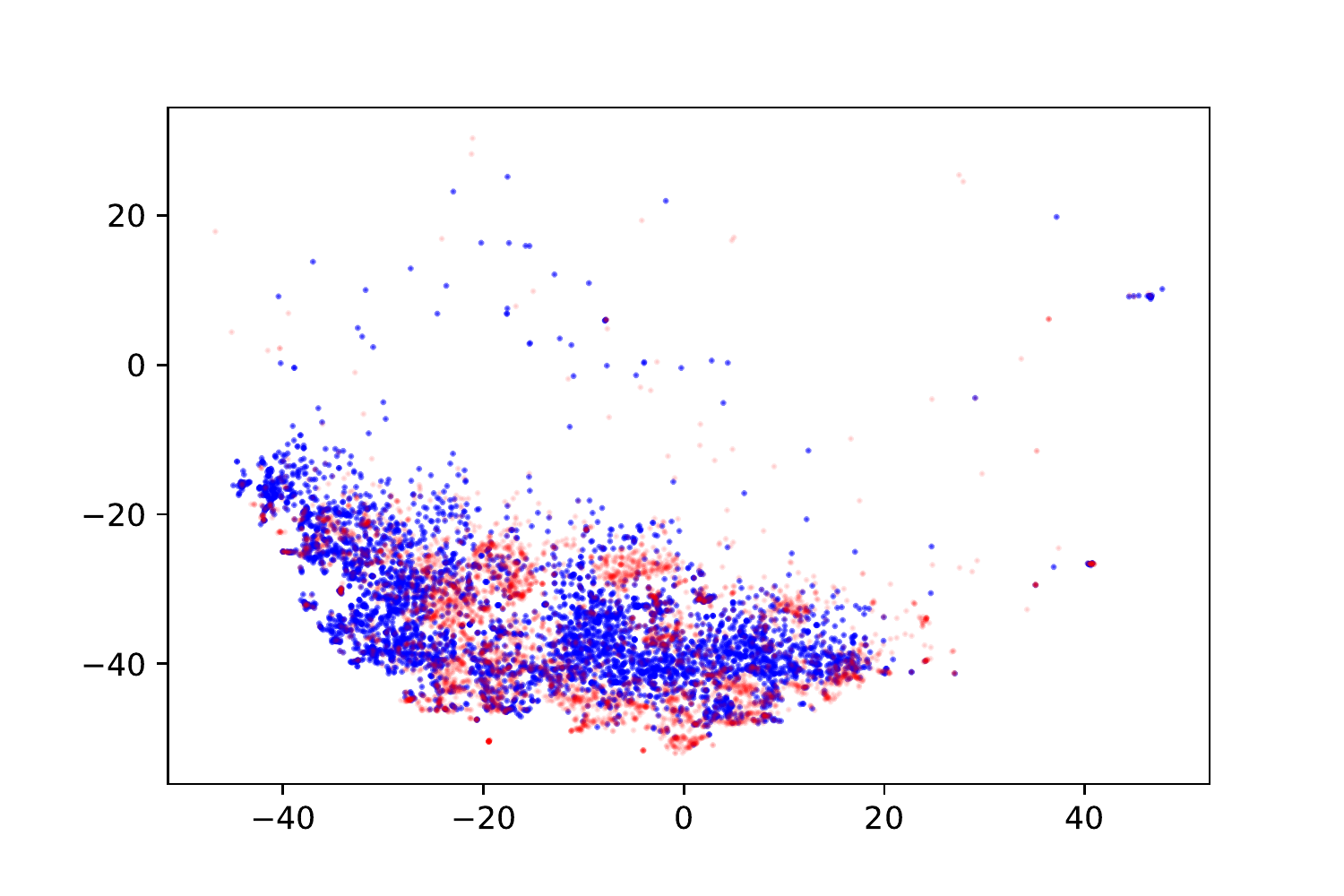}
        \caption{Drama}
        \label{fig:drama}
    \end{subfigure}
    \hspace{10pt}
    \begin{subfigure}{0.3\textwidth}
    \centering
        \includegraphics[width=\textwidth]{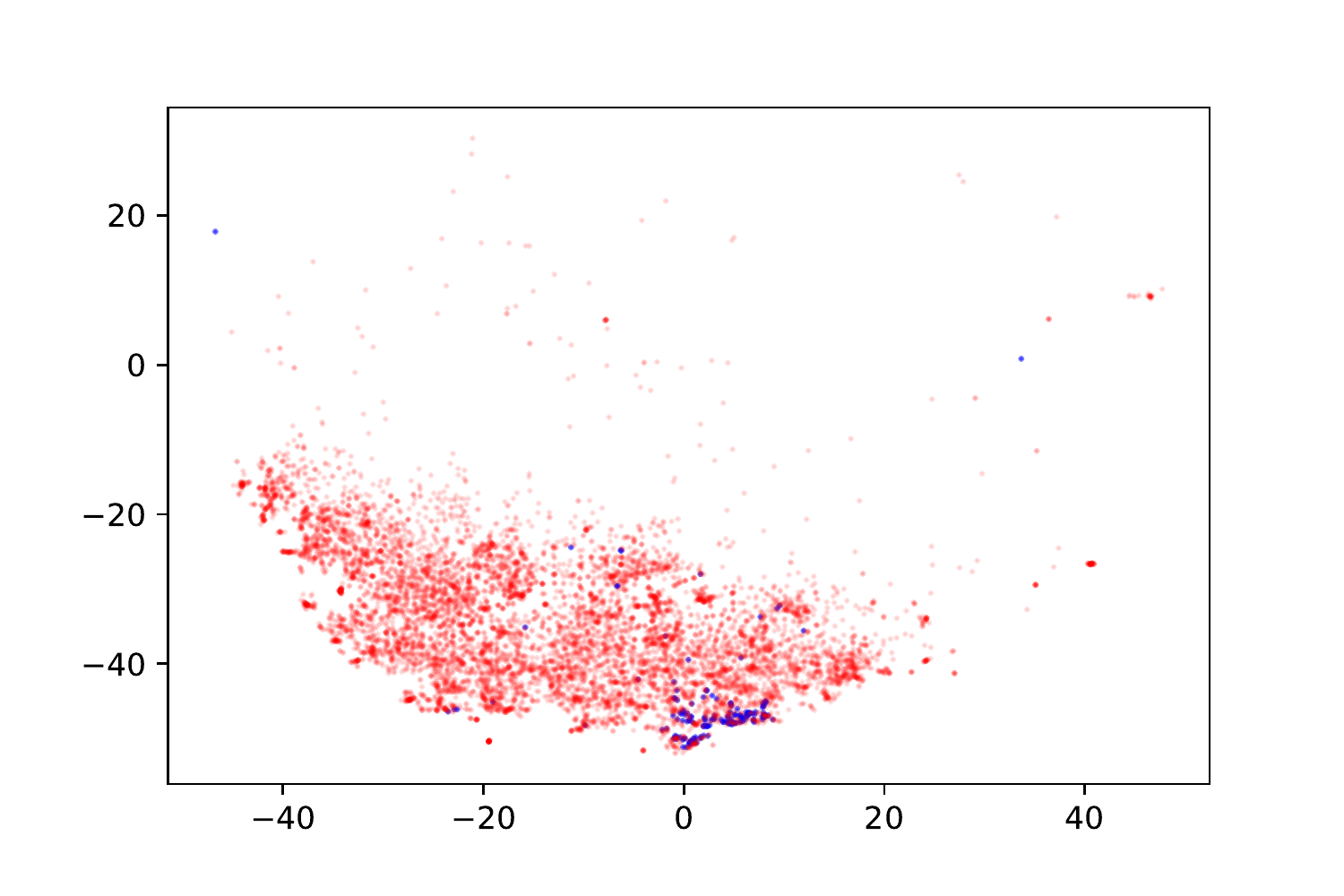}
        \caption{IMAX}
        \label{fig:imax}
    \end{subfigure}
    \vspace{-2pt}
    \vspace{-2pt}
\caption{\textbf{Visualizing movie embeddings in MovieLens 25M by \textit{genres} (Blue dots represent the movies belong to the genre).}}
\vspace{-10pt}
\label{fig:genres}
\end{figure*}

The embeddings can also reflect some genres. In Figure~\ref{fig:genres}, we show the distributions of the following genres: Animation, Children, Horror, the union of Action/Adventure/Sci-fi, Drama, and IMAX. The blue dots in each figure represent movies assigned to a specific genre. We can see some obvious clusters of each genre except for Drama. It is probably because nearly half of the movies are labeled as drama. Another interesting finding of drama movies is their distribution looks very similar to the distribution of the average rating in Figure~\ref{fig:movie_score}, which may be because drama movies that receive more reviews tend to have higher quality, while other movies may not. 


At last, we use three movies to show their nearest neighbors in the 512-dim embedding space. The "distance" between movies is measured by cosine similarity. The top section of Table~\ref{tab:case_study} shows the ten nearest neighbors of "Avengers: Infinity War - Part I (2018)", which is a superhero movie based on a comic series. The ten nearest movies are all superhero movies made by the same media conglomerate, and nine of them are based on same or closely related comic series. The middle section of Table~\ref{tab:case_study} shows the ten nearest movies of "Harry Potter and the Sorcerer's Stone". The list includes all remaining seven movies from the Harry Potter series. The bottom section of Table~\ref{tab:case_study} shows the ten nearest movies of "The Babysitter (2017)," which is a horror comedy. Among its ten nearest movies, seven are labeled as horror, five are labeled as thriller, and two are labeled as comedy. 

 \begin{table*}[!ht]
     \centering
     \caption{Case study: 10 nearest neighbors of movies in embedding space}
     \begin{tabular}{ll}
     \toprule
     Movie    & Genre \\ 
     \midrule
     \multicolumn{2}{c}{10 nearest neighbors of ``\textbf{Avengers: Infinity War - Part I}''} \\
     \midrule
     Avengers: Infinity War - Part II &	Action|Adventure|Sci-Fi \\
     Black Panther & Action|Adventure|Sci-Fi \\
     Thor: Ragnarok & Action|Adventure|Sci-Fi \\
     Guardians of the Galaxy 2 & Action|Adventure|Sci-Fi \\
     Logan &	Action|Sci-Fi \\
     Ant-Man and the Wasp &	Action|Adventure|Comedy|Fantasy|Sci-Fi\\
     Deadpool 2 & Action|Comedy|Sci-Fi \\
     Spider-Man: Into the Spider-Verse &	Action|Adventure|Animation|Sci-Fi \\
     Spider-Man Reboot & Action|Adventure|Fantasy \\
     Incredibles 2 & Action|Adventure|Animation|Children \\
     \midrule
     \multicolumn{2}{c}{10 nearest neighbors of ``\textbf{Harry Potter and the Sorcerer's Stone}''} \\
     \midrule
     Harry Potter and the Chamber of Secrets  & Adventure|Fantasy \\
     Harry Potter and the Prisoner of Azkaban &	Adventure|Fantasy|IMAX \\
     Harry Potter and the Goblet of Fire & Adventure|Fantasy|Thriller|IMAX \\
     Harry Potter and the Order of the Phoenix & Adventure|Drama|Fantasy|IMAX \\
     Harry Potter and the Half-Blood Prince & Adventure|Fantasy|Mystery|Romance|IMAX\\
     Harry Potter and the Deathly Hallows: Part 1 & Action|Adventure|Fantasy|IMAX\\
     Harry Potter and the Deathly Hallows: Part 2 & Action|Adventure|Drama|Fantasy|Mystery|IMAX\\
     Spider-Man  &  Action|Adventure|Sci-Fi|Thriller \\
     Star Wars: Episode II - Attack of the Clones  &	Action|Adventure|Sci-Fi|IMAX \\
     Lord of the Rings: The Fellowship of the Ring & Adventure|Fantasy \\
     \midrule
     \multicolumn{2}{c}{10 nearest neighbors of ``\textbf{The Babysitter}''} \\
     \midrule
     Oculus & Horror \\
     The Voices & Comedy|Crime|Thriller \\
     Unfriended & Horror|Mystery|Thriller \\
     The Visit & Comedy|Horror \\
     Circle & Drama|Horror|Sci-Fi \\
     The Boy & Horror|Thriller\\
     Hush & Thriller \\
     The Shallows & Drama|Thriller \\
     Lights Out & Horror \\
     The Autopsy of Jane Doe & Horror \\ 
     \bottomrule
     \end{tabular}
     \label{tab:case_study}
 \end{table*}

\vskip 1.2in
\section{Related Work}
Recent research work in various fields has demonstrated contrastive learning is an effective self-supervised learning technique, including computer vision~\cite{chen2020simple,radford2021learning}, time series~\cite{zhang2022self}, and recommendation systems~\cite{lin2022improving,zhou2020s3,xie2022contrastive,zhu2021contrastive}.
For example, SimCLR is proposed by~\cite{chen2020simple} as a contrastive learning framework for computer vision. A model pre-trained with SimCLR and fine-tuned with just 1\% of the training data can achieve comparable performance with a model utilizing \textit{all} the available computer vision training data. 
Another example is the CLIP method proposed by~\cite{radford2021learning}, which accomplishes \textit{zero-shot learning} via training a computer vision model jointly with a natural language processing model using contrastive learning techniques. 
Moreover,  the TF-C method proposed by~\cite{zhang2022self} utilizes contrastive learning to transfer knowledge among models trained using time series datasets from various domains. Finally, the NCL work by~\cite{lin2022improving} captures neighborhood information from a bipartite graph with contrastive learning. 
Furthermore, the CL4SRec and COCA methods proposed by \cite{xie2022contrastive} and \cite{zhu2021contrastive} use contrastive learning to extract meaningful information from user behavior sequences. Finally, to enhance the data representations,  \cite{zhou2020s3} proposes the $S^3$-Rec method, where the intrinsic correlation within recommendation datasets is captured by using contrastive learning.

\section{Conclusion}
\label{sec:con} 

This paper studied the problem of learning contextual knowledge from the user-content interaction dataset. 
We first depict the dataset as a bipartite graph. 
Then we demonstrate two crucial and related contexts in the bipartite graph, i.e., the user-side context and the content side context. 
Finally, we frame the goal of learning the knowledge from the two contexts as two contrastive learning tasks. 
We proposed a dual Transformers architecture named PDT to encode the contextual knowledge and show how to apply PDT in the recommendation task. 
Extensive evaluations on two huge popular datasets showed that PDT performs better in 6 metrics than in baselines, including general and sequential recommenders. 
Furthermore, the ablation study showed that both the contextual knowledge extracted from user-side history and content-side history could be beneficial for downstream tasks, and jointly learning the two types of knowledge is the key to the superior performance of PDT. 
In future work, considering applying the PDT to the multivariate sequential data would be an exciting direction. According to our experience with transactional models, we anticipate there are big challenges to deal with large vocabularies and data sparsity issues. The framework can be extended to contrastive learning for more than two aspects. 

\bibliographystyle{plain}
\bibliography{reference}

\end{document}